\documentclass[10pt,twocolumn]{article}

\usepackage[letterpaper,top=0.66in,bottom=0.72in,left=0.66in,right=0.66in,
            columnsep=0.25in,headsep=0.18in]{geometry}
\usepackage{newtxtext,newtxmath}

\usepackage{amsmath,amssymb}
\usepackage{graphicx}
\usepackage{booktabs,tabularx}
\usepackage{microtype}
\usepackage{authblk}
\usepackage{caption}
\usepackage{natbib}
\usepackage[hidelinks]{hyperref}

\hypersetup{
  pdftitle={When Verifiers Vote Backwards under Verdict Substitution: Signed Pivotal Value in Correlated Self-Consistency},
  pdfauthor={Yang Shu}
}
\title{When Verifiers Vote Backwards under Verdict Substitution: Signed Pivotal Value in Correlated Self-Consistency}
\author[1]{Yang Shu\thanks{Corresponding author: \href{mailto:shuyang@zju.edu.cn}{shuyang@zju.edu.cn}.}}
\affil[1]{Zhejiang University, Hangzhou, China}
\date{\small Preprint --- August 2026}

\begin{document}

\maketitle

\begin{abstract}
Replacing one ballot can change a majority decision only on queries decided by a single vote; this structural fact requires no independence assumption. We study the \emph{sign} of that change using a labeled, verdict-style intervention: one correctness signal replaces one correctness-indicator ballot in $k{=}7$ self-consistency panels. This diagnostic intervention is not identical to deployed answer-identity plurality. A primary MATH-500 experiment ($n{=}570$) gives a different-model verifier a $+24.2$pp pivotal gain, whereas a role-reversed configuration gives $-11.2$pp; an exploratory code stress test (14 pivotal rows across 9 tasks) gives $-24.5$pp. An exact signed-gain decomposition accounts for all observed signs through the verifier's state-specific accuracy and the composition of the two one-vote tally states, rather than global accuracy or model provenance. Same-source signals lose accuracy on the pivotal stratum (65$\to$44\% in the primary configuration), while error correlations provide a descriptive error-association diagnostic. Controlled degradation and a $k\in\{3,5,7\}$ subset sensitivity analysis probe the stability of the observed pattern around this accounting. Under the evaluated ties-incorrect answer-identity plurality analysis, the structural zero and strong-verifier benefit persist, but the role-reversed harm attenuates to $-0.9$pp and is not significant. The results therefore establish harmful verdict substitution, not universally harmful deployed plurality, and motivate a testable but unverified hypothesis for negative process-reward-model weights.
\end{abstract}

\section{Introduction}

If you have followed recent work on test-time compute, you may have noticed a strange detail. \citet{kuang2025optimal}, studying how to combine self-consistency votes \citep{wang2023selfconsistency} with process reward model (PRM) scores, report that the Bayes-optimal weight assigned to the PRM is frequently \emph{negative}. This is initially counterintuitive because PRMs are conventionally trained so that larger scores indicate greater correctness. The authors attribute the phenomenon to the failure of a conditional-independence assumption (their Assumption~3.1) but do not explain why a rational optimizer would actively vote \emph{against} its verifier.

For majority-vote panels, only \emph{pivotal} queries -- those decided by one ballot, in the sense of social choice theory's decisive voter \citep{banzhaf1965weighted} -- can change under single-ballot substitution, so gain is exactly zero elsewhere. This arithmetic result does \emph{not} require independent voters. We ask a separate empirical question: once a query is structurally changeable, what determines whether the change is correct or harmful? Self-consistency makes this question acute because its $k$ ballots are samples from one model and share model-specific blind spots.

\begin{figure*}[t]
\centering
\includegraphics[width=0.96\textwidth]{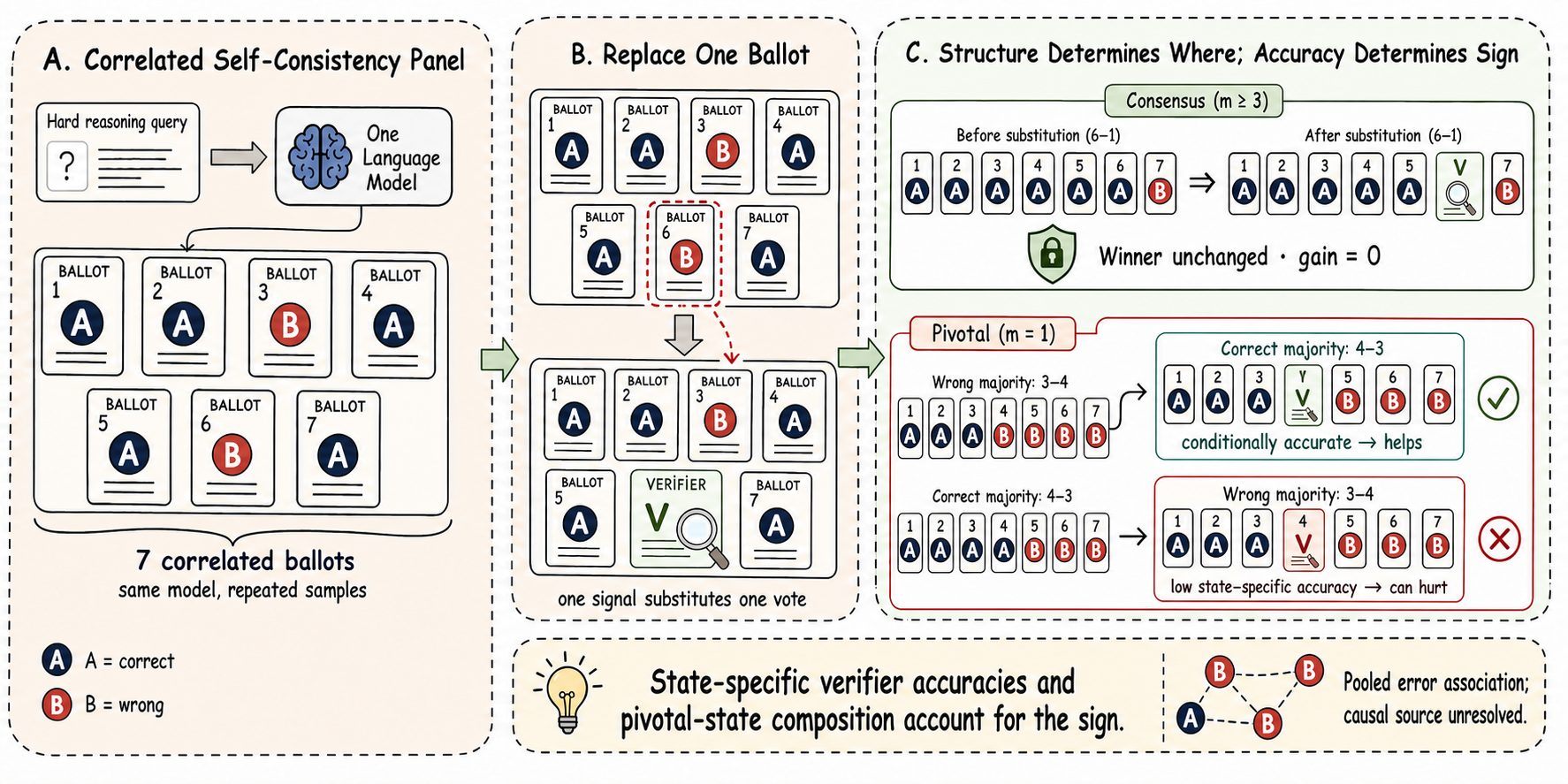}
\caption{Conceptual story of verdict-style single-ballot substitution. (A) Seven correlated samples form a panel; (B) one panel ballot is replaced by a verifier verdict; (C) wider-margin states are structurally inert, whereas signed gain on one-vote states depends jointly on state-specific verifier accuracies and pivotal-state composition. The figure contains no experimental measurements.}
\label{fig:story}
\end{figure*}

Our setting is deliberately narrow: one binary correctness signal replaces one correctness-indicator ballot in a temperature-$0.8$, $k{=}7$ same-model panel (Figure~\ref{fig:story}). Read left to right, the figure separates panel formation, one-ballot replacement, and the two questions that must not be conflated: whether the winner can change and whether a possible change has the correct sign. This is a labeled diagnostic intervention, not a directly deployable aggregator because its ballots require ground truth. We estimate accuracy change separately on one-vote (pivotal) and wider-margin rows, then ask which measured quantities account for its sign; Section~\ref{sec:plurality} separately tests answer-identity plurality.

Our initial conjecture was direct: the pivotal-vote guarantee should survive in self-consistency provided one condition is added -- the verification signal must come from a \emph{different model} than the panel's sampling engine, whereas a same-source signal (such as a PRM trained on the same distribution) should not deliver value. A controlled experiment on hard MATH-500 problems \citep{hendrycks2021measuring,lightman2023lets} appeared to confirm this cleanly: with gpt-4o-mini as the $k{=}7$ self-consistency engine, over the full $190{\times}3$-replication pool ($n{=}570$ rows, zero exclusions), a different-model verifier (DeepSeek-V4-Flash) delivered a $+24.2$pp pivotal-region gain (95\% CI $[14.1, 33.7]$pp), a same-source extra draw delivered exactly $0.0$pp, and both signals' consensus-region gain was exactly zero. Taken alone, this appeared to settle the question.

Two robustness checks, run precisely because the primary result looked clean, overturned it: both came back with the \emph{opposite} sign. A role-reversed configuration -- DeepSeek-V4-Flash as the engine, gpt-4o-mini as the verifier, with independently generated panels on the same 190-problem pool -- produced a pivotal-region gain of $-11.2$pp (two replications, $n{=}380$; both naive and cluster bootstrap CIs exclude zero). Moving to LiveCodeBench code generation \citep{jain2024livecodebench} with execution-based grading produced an exploratory $-24.5$pp estimate. A different-model verifier did not merely fail to help under verdict substitution: in these configurations it flipped pivotal decisions in the wrong direction.

The post-mortem has two layers. The design layer: the standard control -- different-model verifier versus same-source extra draw -- silently confounds provenance with strength, because a same-source draw is accuracy-matched to the panel \emph{in expectation} while the different-model verifier's accuracy is free (ours happened to be 22 points above the panel). The accounting layer organizes all four experiments at once: for binarized $k{=}7$ voting, pivotal gain admits an exact decomposition (Section~\ref{sec:mechanism}) whose inputs are state-specific verifier accuracies and pivotal-state composition. Evaluating the identity on observed counts audits every reported gain (Table~\ref{tab:conditional}), including the case where a globally accuracy-matched verifier was harmful because its accuracy fell from 87\% to 36\% on pivotal queries. Its pooled error correlation with the panel is also higher than the panel's internal correlation; we use this only as a descriptive association because shared task difficulty can contribute to both.

Our contributions: (1) we derive and audit an exact decomposition for \emph{signed} pivotal gain, separating structural changeability from empirical value; (2) across four configurations we document helpful cases, one replicated harmful configuration, and one exploratory harmful accounting case, showing how state-specific accuracy accounts for their observed signs; (3) we manipulate verifier outputs by random flips, report a $k\in\{3,5,7\}$ subset sensitivity, and test five same-source constructions; (4) we show that answer-identity plurality under a ties-incorrect policy preserves the structural zero and strong-verifier benefit but sharply attenuates one negative effect; and (5) we turn these results into a falsifiable hypothesis and a design checklist. Error correlation is reported as a diagnostic association, not a causal explanation. Our claimed novelty is the signed accounting identity and opposite-sign evidence in correlated self-consistency.

\section{Related Work}

\paragraph{Structural support versus signed value.} The elementary changeability fact identifies when a one-ballot substitution can alter majority output; its structural zero is assumption-free. We re-derive that fact in Section~\ref{sec:mechanism}, then study the sign and magnitude of accuracy change in same-model self-consistency panels, including harmful verifiers, conditional decompositions, strength manipulations, math and code domains, and answer-identity voting. Correlated errors among LLM judges are documented at panel level by \citet{kohli2026nine}; multi-judge panels are motivated by \citet{verga2024replacing}.

\paragraph{Correlation ceilings for self-consistency.} \citet{bay2026sampling} generalize the Condorcet jury theorem to correlated sampling, deriving $n_{\mathrm{eff}} = n/(1+(n-1)\rho)$ and asymptotic accuracy under three verifier regimes. Their analysis is aggregate -- total samples versus total accuracy -- and does not stratify by per-query margin. Their ceiling provides theoretical context for why same-source aggregation may fail to improve accuracy; our C7/C17 results remain near single-call performance (Section~\ref{sec:samesource}), making conditional verifier strength, not provenance labels, the binding empirical constraint.

\paragraph{Verifiers that hurt.} \citet{he2026minority} report a verifier baseline with \emph{negative} net effect ($-1.37\%$) in multi-agent debate, attributing it qualitatively to shared blind spots; vote margin appears among their meta-classifier features but is never used for stratified analysis. Our negative configurations provide labeled counterexamples and an exact accounting decomposition for this phenomenon, without identifying its causal origin. \citet{jiang2026discouq} bins multi-agent panels by vote share for uncertainty quantification -- effectively margin stratification -- but with a learned calibration score rather than a literal ballot substitution, so the structural zero does not apply to their setting; the contrast delimits our claim: the exact zero belongs to \emph{single-ballot replacement}, not to margin stratification per se.

\paragraph{Answer identities versus correctness indicators.} \citet{cai2026arbiter} study pure self-consistency through reasoning-trajectory basins over \emph{answer identities}, with deliberately zero external information. Their object of study highlights a limitation of correctness-indicator analyses like our main one: deployed self-consistency takes a plurality over answer identities, and collapsing all wrong answers into one class can change which queries are contested. We confront this directly rather than in passing: Section~\ref{sec:plurality} re-analyzes our math experiments under answer-identity plurality voting -- possible at zero cost because every sampled answer string is banked -- and reports which conclusions strengthen, survive, or attenuate.

\paragraph{PRM aggregation.} \citet{kuang2025optimal} derive optimal weights for combining self-consistency with PRM scores under conditional independence (their Theorem~3.2) and observe empirically that optimal PRM weights are often negative, without a mechanism. We propose one as a hypothesis (Section~\ref{sec:discussion}) and state what would falsify it.

\section{Structural Support Versus Empirical Sign}
\label{sec:mechanism}

\paragraph{Setup and notation.} A \emph{panel} is $k{=}7$ separately generated temperature-$0.8$ samples of one model (the \emph{engine}) on a query. In the \emph{binarized} scheme used by our main analyses, each sample's correctness indicator is a vote; the panel decision is the majority; with $S_i$ correct votes on query $i$ the \emph{margin} is $m_i = |2S_i - k|$, $m_i{=}1$ is \emph{pivotal} ($S_i \in \{3,4\}$), $m_i \geq 3$ is \emph{consensus}. \emph{Panel per-call accuracy} means the average correctness of individual samples (not of the majority decision). A \emph{signal} is one additional call whose binarized verdict $V$ replaces one panel vote; we evaluate leave-one-out over all 7 positions and report the accuracy change per stratum, with $10{,}000$-iteration bootstrap CIs, row-resampled (naive) and task-clustered. The deployed \emph{answer-identity} scheme is analyzed in Section~\ref{sec:plurality}.

\paragraph{Structural support.} Replacing one ballot changes the vote total by at most one, so it can flip the majority only when $m_i{=}1$. This is arithmetic, not statistics: it invokes no independence assumption. The prediction is that the consensus-region gain is \emph{exactly} zero -- not small, zero -- regardless of correlation, signal quality, or effect direction elsewhere. All 13 signal--experiment combinations below satisfy it to floating-point precision, as they must. In view of the concurrent result, we use this as an implementation audit rather than claim it as a new finding.

\paragraph{Signed value.} Whether a pivotal vote is flipped in the \emph{right} direction is an empirical matter with an exact answer. Averaging the leave-one-out substitution over the 7 drop positions,
\begin{align}
\label{eq:decomp}
G_{\mathrm{piv}} = \tfrac{4}{7}\Big[& P(S{=}3 \mid m{=}1)\, P(V{=}1 \mid S{=}3) \nonumber\\
 -\ & P(S{=}4 \mid m{=}1)\, \big(1 - P(V{=}1 \mid S{=}4)\big)\Big],
\end{align}
because with $S{=}3$ the substitution helps iff a wrong vote is dropped (probability $4/7$) and $V{=}1$, and with $S{=}4$ it hurts iff a right vote is dropped (probability $4/7$) and $V{=}0$. The operative inputs are therefore the verifier's two \emph{state-specific pivotal accuracies}, $P(V{=}1\mid S{=}3)$ and $P(V{=}1\mid S{=}4)$, together with pivotal-state composition -- not a single global accuracy. The global verifier-minus-panel gap is a proxy, and Section~\ref{sec:results} shows both when it works and how it fails: it fails for signals whose accuracy \emph{redistributes} across strata; pooled error correlation is reported only as a descriptive signature of where such redistribution may occur.

\paragraph{The trap in the standard control.} The natural experiment -- different-model verifier versus same-source extra draw -- appears to isolate provenance. It does not. The same-source arm is accuracy-matched to the panel in expectation (it \emph{is} the panel's model), while the different-model arm's accuracy is free, so any observed advantage can be explained by capability rather than decorrelation. Worse, the same-source arm shares the panel's model and sampling distribution; in our experiments its \emph{conditional} accuracy falls sharply on the pivotal queries selected by panel disagreement. Separating the factors requires configurations where the different-model verifier holds no accuracy advantage (our two negative checks), direct strength manipulation at fixed provenance (Section~\ref{sec:degradation}), and attempts to hand the advantage to the same-source arm (Section~\ref{sec:samesource}).

\section{Experimental Setup}

\paragraph{Domains, grading, completeness.} \emph{Math}: using seed 42, we sampled 190 MATH-500 level-4/5 problems whose reference answers pass the project's simple-numeric filter \citep{hendrycks2021measuring,lightman2023lets}; grading is exact match after final-answer normalization. This restriction makes answer extraction reliable but narrows external validity beyond difficulty level alone. \emph{Code}: 60 LiveCodeBench problems \citep{jain2024livecodebench}, graded by sandboxed execution of held-out tests. Rows are (problem $\times$ replication) pairs; every replication independently re-runs the full protocol. Completion accounting, per configuration: primary $570/570$ rows ($190 \times 3$, zero exclusions), swap $380/380$ ($190 \times 2$, zero exclusions), LCB strong-verifier $240/240$ ($60 \times 4$, zero exclusions), LCB original Signal A $248/300$ ($60 \times 5$ planned; 52 rows lost to non-extractable generations in early replications, before the escalation protocol below was adopted). One of the 248 rows lacks the same-source Signal B output, so analyses requiring both signals use 247; that row is consensus and cannot affect pivotal estimates. An earlier version of the primary analysis used $526/570$ rows because 44 had a missing verifier verdict; those rows are \emph{not} missing at random (their panels are harder), so we completed the pool and report full-pool numbers throughout -- doing so \emph{raised} the primary pivotal gain from $+23.3$ to $+24.2$pp.

\paragraph{Configurations.} Table~\ref{tab:main} lists the four main experiments. The primary configuration uses gpt-4o-mini as engine and DeepSeek-V4-Flash as verifier; the role-reversed check exchanges model roles but generates independent panels on the same task pool. The code configurations use (Kimi-K2.7-Code, gpt-4.1-mini) and (doubao-1-5-lite, gpt-5.4), the latter chosen deliberately for a large accuracy gap. Every experiment also carries a same-source Signal~B (an extra draw of the engine).

\paragraph{Token-budget escalation.} DeepSeek-V4-Flash, a reasoning model, exhausts a 4{,}096-token budget on the hardest problems before emitting visible text. A single-variable probe confirmed truncation, not refusal: the same problem answers normally at 16{,}384 tokens. We therefore use a uniform protocol -- sample at 4{,}096; on an \emph{empty} output (which carries no vote) redraw at 16{,}384, then 32{,}768. Escalation replaces only non-votes, but it does give the hardest problems more compute; as a sensitivity check, the swap configuration restricted to the 165-problem subset collectable without escalation gives $-9.9$pp (CI crossing zero) versus $-11.2$pp (significant) on the full pool -- same direction, sharpened by completing the hard problems.

\begin{table*}[t]
\centering
\caption{Four main experiments with different-model verifiers. Gap and gain are absolute percentage-point differences; consensus gain is zero in every row.}
\label{tab:main}
\small
\setlength{\tabcolsep}{5pt}
\begin{tabular}{@{}llrrrrl@{}}
\toprule
Experiment & Engine / Verifier & Gap (pp) & $n$ (pivotal) & Pivotal gain & 95\% CI (naive) & Consensus \\
\midrule
Primary (MATH-500) & 4o-mini 64.1 / DS-V4-Flash 86.1 & $+22.0$ & 570 (59) & $+24.2$pp & $[+14.1, +33.7]$ & $0.0000$ \\
Role-reversed (MATH-500) & DS-V4-Flash 84.1 / 4o-mini 63.2 & $-20.9$ & 380 (51) & $-11.2$pp & $[-20.9, -1.2]$ & $0.0000$ \\
Original config (LCB, code) & K2.7-Code 87.9 / 4.1-mini 87.1 & $-0.8$ & 248 (14) & $-24.5$pp & $[-40.8, -9.5]$ & $0.0000$ \\
Strong-verifier (LCB, code) & doubao-lite 68.3 / gpt-5.4 98.8 & $+30.5$ & 240 (36) & $+17.5$pp & $[+9.2, +26.5]$ & $0.0000$ \\
\bottomrule
\end{tabular}
\end{table*}

Task-clustered CIs agree with the naive CIs in sign: primary $[+12.2,+35.5]$, role-reversed $[-21.1,-1.1]$, LCB original $[-42.9,-5.7]$, and LCB strong $[+9.8,+26.4]$pp. The LCB-original estimate is exploratory: its 14 pivotal rows come from only 9 unique tasks. Same-source Signal~B is non-significant in every configuration.

\section{Results}
\label{sec:results}

\subsection{Evidence Across Four Configurations}

\begin{figure}[t]
\centering
\includegraphics[width=\columnwidth]{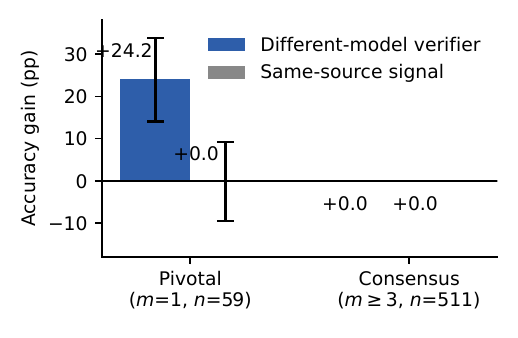}
\caption{Primary MATH-500 gains by margin stratum ($n{=}570$). Error bars are naive-bootstrap 95\% CIs.}
\label{fig:bars}
\end{figure}

\paragraph{Primary result.} Over the full pool ($n{=}570$; 59 pivotal), the different-model verifier gains $+24.2$pp in the pivotal region, the same-source extra draw gains exactly $0.0$pp (CI $[-9.4, +9.3]$), and both consensus-region gains are exactly zero (Figure~\ref{fig:bars}). Taken alone, this reads as a clean confirmation that provenance is the active ingredient.

\paragraph{Role-reversed result.} Exchanging the two models' roles while generating fresh panels on the same pool inverts the effect: $-11.2$pp over $n{=}380$ rows and 51 pivotal cases; both unadjusted 95\% bootstrap CIs exclude zero. Because configurations were selected adaptively and the exploratory family is not multiplicity-adjusted, we treat this as a replicated negative configuration rather than a confirmatory population test.

\paragraph{Exploratory code stress test.} The original code configuration -- verifier globally accuracy-matched to the panel ($-0.8$pp) -- yields $-24.5$pp (14 pivotal rows across 9 tasks; naive 95\% CI $[-40.8,-9.5]$, cluster $[-42.9,-5.7]$). Its effective sample is too small for a broad population claim; we use it only to show that global matching need not imply neutrality and to motivate the accounting in Section~\ref{sec:conditional}.

\paragraph{Targeted positive check.} The deliberately selected strong-verifier configuration ($+30.5$pp gap, different domain, different model pair) restores $+17.5$pp over four replications ($n{=}240$, 36 pivotal). The point estimate relaxed from $+25.0$pp at two replications while the CI tightened; this supports the narrowed accounting picture but is not a prospectively registered confirmation.

\subsection{Exact Signed-Gain Decomposition}
\label{sec:conditional}

Table~\ref{tab:conditional} reports every empirical input needed to evaluate Eq.~\ref{eq:decomp}. The calculated and observed gains agree because the equation is an accounting identity; the agreement audits code and bookkeeping rather than validating a predictive or causal model. DeepSeek-V4-Flash keeps its accuracy on pivotal queries (86.1\% global $\to$ 86.4\% pivotal), whereas the same-source draw falls from 65.4\% to 44.1\%. In the exploratory LCB-original case, gpt-4.1-mini falls from 87.0\% to 35.7\%, including $0/3$ correct verdicts in the small $S{=}3$ cell and $5/11$ in the $S{=}4$ cell. Its pooled error correlation with individual panel ballots is $0.53$, above the panel's internal $0.46$; helpful verifiers have $0.07$ and $-0.03$. These associations flag where to inspect conditional performance but can reflect shared task difficulty and do not identify a causal dependence mechanism.

\begin{table*}[t]
\centering
\caption{Observed inputs and accounting outputs for Eq.~\ref{eq:decomp}. Here $q_4=P(S{=}4\mid m{=}1)$, $a_s=P(V{=}1\mid S{=}s)$, and $\varphi_V/\varphi_P$ compares signal--ballot with within-panel error correlation. The $S{=}3/S{=}4$ row counts are 33/26 (primary), 13/38 (role-reversed), 3/11 (LCB original), and 12/24 (LCB strong).}
\label{tab:conditional}
\small
\setlength{\tabcolsep}{4.4pt}
\begin{tabular}{@{}lrrrrrrr@{}}
\toprule
Config / signal & $n_{\rm piv}$ & $q_4$ & $a_3$ & $a_4$ & $V$ acc.\ g$\to$p (\%) & $G_{\rm calc}/G_{\rm obs}$ (pp) & $\varphi_V/\varphi_P$ \\
\midrule
Primary / DS-V4-F & 59 & .441 & .939 & .769 & 86.1$\to$86.4 & $+24.2/+24.2$ & .07/.66 \\
Primary / same-src & 59 & .441 & .364 & .538 & 65.4$\to$44.1 & $0.0/0.0$ & .63/.66 \\
Role-rev.\ / 4o-mini & 51 & .745 & .462 & .579 & 63.2$\to$54.9 & $-11.2/-11.2$ & .07/.13 \\
Role-rev.\ / same-src & 51 & .745 & .769 & .816 & 83.4$\to$80.4 & $+3.4/+3.4$ & .07/.13 \\
LCB-orig / 4.1-mini & 14 & .786 & .000 & .455 & 87.0$\to$35.7 & $-24.5/-24.5$ & .53/.46 \\
LCB-orig / same-src & 14 & .786 & .000 & .727 & 88.3$\to$57.1 & $-12.2/-12.2$ & .46/.46 \\
LCB-strong / gpt-5.4 & 36 & .667 & .917 & 1.000 & 98.8$\to$97.2 & $+17.5/+17.5$ & $-.03/.60$ \\
LCB-strong / same-src & 36 & .667 & .417 & .792 & 67.9$\to$66.7 & $0.0/0.0$ & .59/.60 \\
\bottomrule
\end{tabular}
\end{table*}

\subsection{Attempts to Strengthen Same-Source Signals}
\label{sec:samesource}

The narrowed picture leaves one corner untested: a \emph{same-source} signal genuinely more accurate than the panel. We tried both zero-training routes; in our configurations neither produced a substantial, reliable advantage (Table~\ref{tab:samesource}).

\paragraph{Route 1: aggregation.} For each row we aggregate same-source draws \emph{outside} its own panel into one ballot: in math, the majority of another replication's 7 draws (C7) or of all 17 other banked draws (C17); in code, C7 and C25. Under independence, majority-of-17 should far exceed a single call; measured, C7 $= 64.0\%$ and C17 $= 63.9\%$ remain near the single-call $64.1\%$, consistent with the correlation ceiling of \citet{bay2026sampling}. In code, aggregation buys $+4.6$ and $+5.5$pp, but the corresponding pivotal gains remain small and non-significant.

\paragraph{Route 2: strengthened elicitation.} Same weights (gpt-4o-mini), meticulous-verifier prompt, temperature $0.2$: $63.5\%$ per-call against the panel's $64.3\%$ -- slightly below. Pivotal gain $+3.9$pp, non-significant; consensus exactly zero.

\begin{table}[t]
\centering
\caption{Five same-source signal constructions. None yields a statistically reliable pivotal gain.}
\label{tab:samesource}
\small
\setlength{\tabcolsep}{3.5pt}
\begin{tabular}{@{}lrrr@{}}
\toprule
Signal & Acc.\ vs.\ panel & Gap (pp) & Pivotal gain \\
\midrule
Math C7 (maj.\ of 7) & 64.0 / 64.1 & $-0.1$ & $+1.9$pp (ns) \\
Math C17 (maj.\ of 17) & 63.9 / 64.1 & $-0.2$ & $-4.8$pp (ns) \\
Code C7 (maj.\ of 7) & 72.9 / 68.3 & $+4.6$ & $+3.2$pp (ns) \\
Code C25 (maj.\ of 25) & 73.8 / 68.3 & $+5.5$ & $+4.8$pp (ns) \\
Verifier prompt, $T{=}0.2$ & 63.5 / 64.3 & $-0.8$ & $+3.9$pp (ns) \\
\bottomrule
\end{tabular}
\end{table}
\paragraph{Consequence.} In these five constructions, resampling and prompting did not give a same-source signal a reliable advantage. Stronger interventions such as supervised training remain untested. Eq.~\ref{eq:decomp} makes a falsifiable prediction for any such intervention: positive pivotal value requires sufficient accuracy in the two pivotal tally states, weighted by their observed composition.

\subsection{Manipulating Strength Directly}
\label{sec:degradation}

\begin{figure}[t]
\centering
\includegraphics[width=\columnwidth]{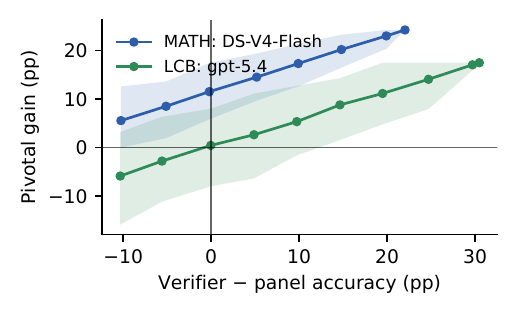}
\caption{Asymmetric oracle degradation that flips correct verifier outputs to incorrect ones. Shading shows the 2.5--97.5 percentiles over 200 flip seeds, not task-sampling uncertainty.}
\label{fig:degradation}
\end{figure}

The four main experiments vary provenance, task, and strength together. To manipulate strength while holding the observed panel, task set, prompts, and model pair fixed, we degrade a strong verifier by randomly flipping correct verdicts to incorrect until a target accuracy is reached (200 flip seeds per target). Figure~\ref{fig:degradation}: in both domains the pivotal gain falls smoothly and monotonically under this one-sided corruption, crossing into negative territory at negative gaps in the code configuration ($-5.9$pp at a $-10$pp gap). Because the corruption is sampled independently of panel outcomes, it avoids deliberately targeting pivotal queries; at a globally matched accuracy the degraded verifier still gains $+11.5$pp, whereas the equally matched but panel-correlated organic verifier of the LCB original configuration loses $-24.5$pp. The intervention identifies the effect of this corruption in the fixed observed configurations, not a general causal effect of verifier strength or dependence; the contrasts are organized descriptively by the conditional pivotal-state accuracies in Eq.~\ref{eq:decomp}.

\paragraph{Panel-size subset sensitivity.} Using the three complete primary replications, we enumerate every $k$-subset of each banked seven-sample panel and task-cluster bootstrap the induced estimand. For the different-model verifier, pivotal gains are $+23.0$pp at $k{=}3$ (CI $[+17.3,+28.4]$), $+21.2$pp at $k{=}5$ ($[+14.2,+27.8]$), and $+24.2$pp at $k{=}7$ ($[+12.3,+35.0]$). Same-source estimates are $+3.0$, $+0.7$, and $0.0$pp, with every CI crossing zero. Because smaller panels reuse subsets of the banked $k{=}7$ draws rather than independent generations, this check supports stability within the primary sample but does not test larger panels or new sampling trajectories.

\subsection{Robustness: Deployed Answer-Identity Voting}
\label{sec:plurality}

Deployed self-consistency takes a plurality over \emph{answer identities}, with no ground-truth access at aggregation time; our main analyses instead binarize votes into correctness indicators. This distinction sharply bounds the practical negative claim. Because every sampled answer string is banked, the math experiments can be re-analyzed under the deployed scheme at zero cost: panel = 7 answer strings, decision = plurality winner (ties graded incorrect), margin = top-count minus runner-up count, and the verifier substitutes its own \emph{answer}, not a verdict. The structural guarantee transfers with the same threshold: one substitution moves any answer's count by at most one down and one up, so the winner cannot change when the plurality margin is $\geq 3$ -- and empirically the consensus-region gain is exactly zero in all four signal--configuration pairs (bootstrap CIs literally $[0,0]$). On the reachable stratum (margin $\leq 2$; $n{=}158$ of 570 rows), the strong verifier's gain remains large and significant: $+15.0$pp (naive CI $[10.4, 19.6]$, cluster $[9.8, 20.5]$). Two honest attenuations: the same-source extra draw now gains a small but significant $+4.5$pp (CI $[0.9, 8.3]$), while the swap configuration's harm shrinks from $-11.2$pp to $-0.9$pp (ns). A fresh draw is biased toward the engine's modal answer relative to the scattered ballots that pivotal conditioning selects; likewise, a weak verifier's wrong \emph{answer} is usually a novel identity that cannot dethrone the mode, whereas a wrong \emph{verdict} counts directly against the correct side. ``Voting backwards'' at full strength is therefore established for verdict-style substitution, not as a universal property of deployed plurality. The strong-verifier benefit and consensus zero survive both schemes. Code-domain plurality requires behavioral clustering of programs, which our banked data do not support (Section~\ref{sec:limitations}).

\subsection{Thirteen Points, One Association}

\begin{figure*}[t]
\centering
\includegraphics[width=0.92\textwidth]{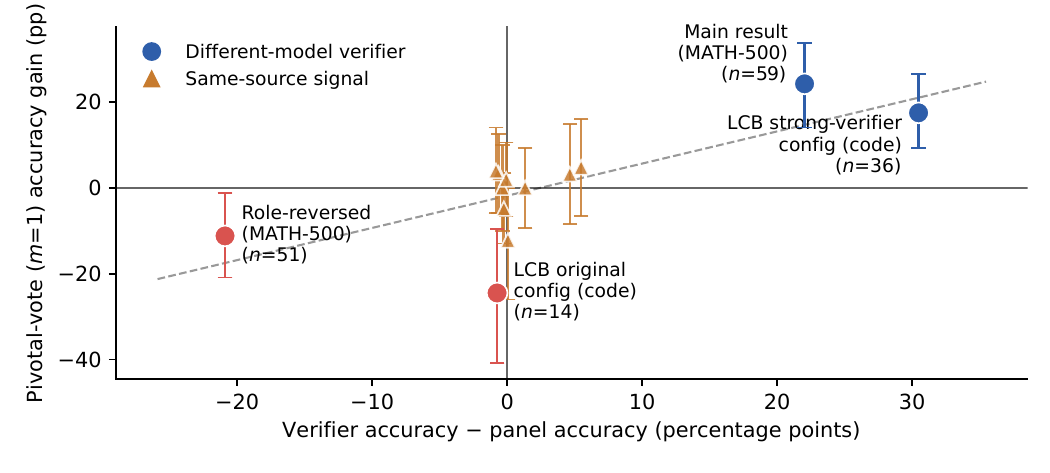}
\caption{Pivotal gain versus global verifier--panel accuracy gap for 13 non-independent combinations. Error bars are naive-bootstrap 95\% CIs; marker shape denotes provenance and the dashed OLS line is descriptive only.}
\label{fig:gap}
\end{figure*}

Figure~\ref{fig:gap} assembles every binarized signal--experiment combination: four different-model verifiers and nine same-source signals across two domains and three constructions. The dashed line is a descriptive visual aid only: the 13 points share panels, are non-independent, and support no regression inference. The coarse trend is useful precisely because its large residual (LCB original, the red point at near-zero gap) is the signal whose accuracy does not survive pivotal conditioning.

\section{Discussion}
\label{sec:discussion}

\paragraph{A hypothesis for why optimal PRM weights go negative.} Chaining the findings suggests a testable account, stated as a hypothesis -- we evaluate no actual PRM, and weighted score aggregation is not identical to single-ballot substitution. A PRM may behave as a dependence-bearing, same-source-like signal when its errors align with those of the engine it scores; this is not guaranteed by the label ``PRM.'' In our experiments, same-source signals did not exceed their panel's accuracy without injected training information, and their conditional accuracy collapsed on pivotal queries. If these regularities carry over, an optimizer combining votes and PRM scores gains little from treating the PRM as added evidence exactly where evidence matters, but \emph{can} profit from using it to cancel bias it shares with the panel -- a negative weight. Two tests would falsify or confirm this: evaluating an actually-trained PRM's conditional pivotal-state accuracies, and bridging ballot substitution to weighted aggregation formally. The regularity of \citet{kuang2025optimal} is consistent with this account but not established by our experiments.

\paragraph{A design checklist for verifier-value experiments.} (1) Report every signal's accuracy relative to the panel, globally \emph{and in each pivotal tally state} -- the ``extra draw of the same model'' control is accuracy-matched globally by construction and collapsed pivotally in our experiments, so it cannot separate provenance from strength. (2) Measure dependence ($\varphi$ or similar) instead of assuming it from model identity, while recognizing that pooled association also reflects task difficulty. (3) Include at least one reversed configuration where the different-model signal holds no accuracy advantage. Had we not run ours, this paper would have reported a clean, replicated result whose unadjusted CIs excluded zero, yet whose provenance interpretation was overturned by two further experiments.

\paragraph{The structural zero as an implementation audit.} The consensus-region zero held to floating-point precision in every configuration, under both voting schemes. It is an arithmetic identity, and we present it as such: any single-ballot-substitution analysis inherits it, so deviation from exact zero indicates a bug, not an effect.

\section{Limitations}
\label{sec:limitations}

Our main panels use $k{=}7$ and one temperature per role. The $k{=}3,5$ analysis reuses subsets of those draws; independently generated smaller panels, larger $k$, and a systematic temperature sweep remain untested. The 13-point gap--gain plot is descriptive, and its shared panels preclude regression inference; configuration-wise intervals are not multiplicity-adjusted for the exploratory family. Role-reversed panels are independent draws, so their realized pivotal subsets are not paired. The LCB original configuration has only 14 pivotal rows across 9 tasks and 52/300 rows lost to early-protocol extraction failures; one additional consensus row lacks Signal B. We therefore treat it as an exploratory accounting case, not evidence for a population effect. Pooled $\varphi$ can reflect shared task difficulty and is not residualized. Random-flip intervals cover corruption seeds, not task sampling, and the intervention only corrupts correct outputs. The answer-identity re-analysis covers math only and uses one answer normalizer and a conservative ``ties incorrect'' policy; alternative semantic-equivalence and tie policies are not tested. Our code pipeline banked pass/fail verdicts but not program text, so behavioral clustering is impossible retroactively. An actual PRM and weighted score aggregation remain untested; the PRM account is consequently a hypothesis. Finally, escalated token budgets give the hardest problems more compute; the fixed-budget sensitivity result in Section~4 bounds the effect in the one configuration it touches.

\section{Reproducibility and Ethics}

The research artifacts include raw per-call samples, analysis scripts, prompts, model IDs, call-date ranges, sampling parameters, bootstrap seeds (777 naive / 778 cluster / 20260727 degradation / 20260728 panel-size), and a script-to-table reproduction map. The three data figures (Figures~\ref{fig:bars}--\ref{fig:gap}) and every reported number regenerate from banked samples. Figure~\ref{fig:story} is an AI-assisted conceptual schematic selected and checked by the author; it encodes no experimental measurements, and the final source PNG is included. The work uses public benchmarks and commercial LLM APIs; no human subjects are involved. We see no direct ethical risks beyond the general dual-use considerations of improving LLM evaluation.

\section{Conclusion}

Pivotal structure determines where one-ballot substitution can matter, but not whether it helps. Under verdict-style correctness-indicator substitution, an exact accounting identity shows that gain depends on verifier accuracy in the two one-vote tally states and on their composition; global accuracy and provenance can mislead. Across our configurations, several same-source signals lose accuracy on those queries, one different-model verifier is harmful under verdict substitution, and controlled degradation traces what happens under asymmetric one-sided corruption. Under the evaluated ties-incorrect answer-identity analysis, plurality preserves the structural zero and strong-verifier benefit but attenuates the role-reversed harm to a non-significant estimate, so harmful deployed plurality is not established. The account describes observed signs without claiming that correlation causes them. Testing actual PRMs, weighted aggregation, independently generated panel sizes, and broader answer-identity settings is necessary to evaluate the proposed negative-weight hypothesis.

\bibliographystyle{plainnat}
\bibliography{references}

\end{document}